\documentclass[10pt,twocolumn,letterpaper]{article}
\usepackage[pagenumbers]{cvpr}    
\usepackage[dvipsnames]{xcolor}

\definecolor{mygreen}{RGB}{112,173,71}
\definecolor{myblue}{RGB}{91,155,213}
\definecolor{myorange}{RGB}{237,125,49}
\definecolor{myred}{RGB}{255,22,67}
\definecolor{honey}{RGB}{201,89,95}
\definecolor{amber}{RGB}{180,148,41}
\definecolor{darkblue}{RGB}{84,96,126}

\newif\ifdraft
\drafttrue
\ifdraft
\newcommand{\hwdc}[1]{{\color{blue}[\textbf{Howard:} #1]}}
\newcommand{\kac}[1]{{\color{purple}[\textbf{Kfir:} #1]}}
\newcommand{\yac}[1]{{\color{red}[\textbf{Yuval:} #1]}}
\newcommand{\jwc}[1]{{\color{orange}[\textbf{Jian} #1]}}

\newcommand{\todo}[1]{{\color{blue}[TODO: #1]}}

\else
\newcommand{\hwdc}[1]{}
\newcommand{\kac}[1]{}
\newcommand{\yac}[1]{}
\newcommand{\jwc}[1]{}
\newcommand{\todo}[1]{}

\fi

\usepackage{graphicx}
\usepackage{amsmath}
\usepackage{amssymb}
\usepackage{booktabs}
\usepackage[dvipsnames]{xcolor}
\usepackage{tabularx, booktabs}
\usepackage{multirow}
\usepackage{colortbl}

\definecolor{cvprblue}{rgb}{0.21,0.49,0.74}
\usepackage[pagebackref,breaklinks,colorlinks,allcolors=cvprblue]{hyperref}

\newcommand{\sr}{SiR\xspace}

\newcommand{\textmethod}{structured text prediction\xspace}

\newcommand{\TextMethod}{Structured Text Prediction\xspace}
\newcommand{\groundedmethod}{grounded prediction\xspace}
\newcommand{\GroundedMethod}{Grounded Prediction\xspace}

\newcommand{\attentionmethod}{attention feature augmentation\xspace}

\usepackage{listings}

\newbox\jsavebox
\newcommand{\jsubfig}[2]{%
	\sbox\jsavebox{#1}%
	\parbox[t]{\wd\jsavebox}{\centering\usebox\jsavebox\\#2}%
	}

\usepackage[capitalize]{cleveref}
\crefname{section}{Sec.}{Secs.}
\Crefname{section}{Section}{Sections}
\Crefname{table}{Table}{Tables}
\crefname{table}{Tab.}{Tabs.}

\def\paperID{4151} %
\def\confName{CVPR}
\def\confYear{2025}

\title{Dynamic Scene Understanding from Vision-Language Representations}

\author{{\centering Shahaf Pruss$^{*1}$ \: Morris Alper$^{*1}$ \:  \: Hadar Averbuch-Elor$^{1,2}$ 
        }
\\
{\parbox{0.9\textwidth}{\centering
$^1$Tel Aviv University \quad
        $^2$Cornell University  
       }
}}

\begin{document}

\maketitle
\def\thefootnote{*}\footnotetext{These authors contributed equally to this work}\def\thefootnote{\arabic{footnote}}

\begin{abstract}
    Images depicting complex, dynamic scenes are challenging to parse automatically, requiring both high-level comprehension of the overall situation and fine-grained identification of participating entities and their interactions. Current approaches use distinct methods tailored to sub-tasks such as Situation Recognition and detection of Human-Human and Human-Object Interactions.
However, recent advances in image understanding have often leveraged web-scale vision-language (V\&L) representations to obviate task-specific engineering.
In this work, we propose a framework for dynamic scene understanding tasks by leveraging knowledge from modern, frozen V\&L representations.
By framing these tasks in a generic manner --- as predicting and parsing structured text, or by directly concatenating representations to the input of existing models -- we achieve state-of-the-art results while using a minimal number of trainable parameters relative to existing approaches.
Moreover, our analysis of dynamic knowledge of these representations shows that recent, more powerful representations effectively encode dynamic scene semantics, making this approach newly possible. Project page: \href{https://tau-vailab.github.io/Dynamic-Scene-Understanding/}{https://tau-vailab.github.io/Dynamic-Scene-Understanding/}.

\end{abstract}

\begin{figure}[t]
\center
    \includegraphics[width=0.5\textwidth]{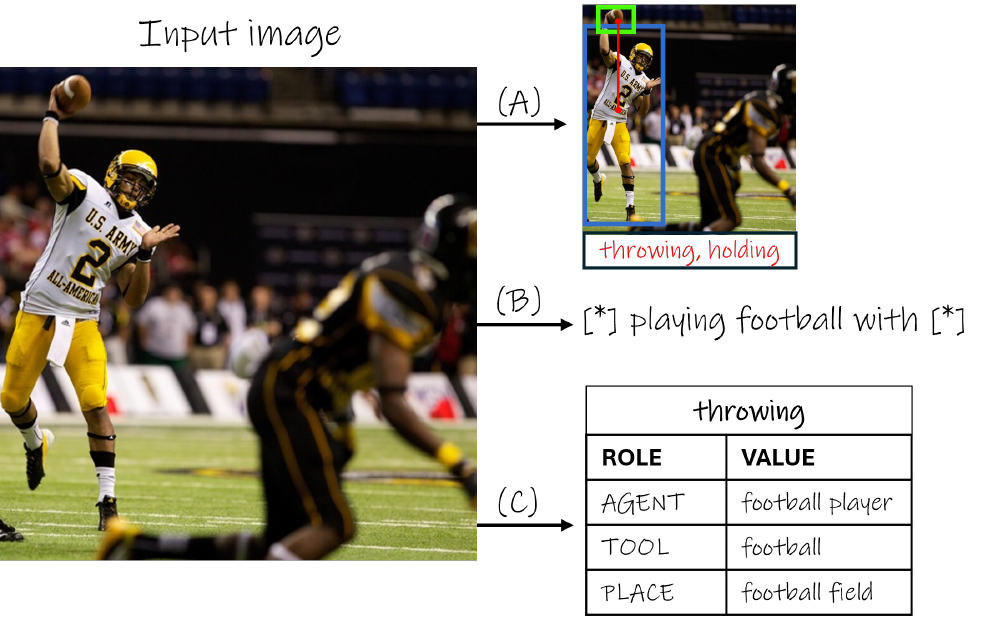}
    \caption{
    Given an input image depicting a dynamic scene (left),
    our framework performs a variety of dynamic scene understanding tasks, such as human-object interactions, human-human and recognition of grounded situations, (A, B, C respectively above). Each of these predicts different entities and relations, possibly grounded in the input image (visualized as bounding boxes on the right). Our generic method contrasts with previous approaches tailored to a single such task.
    }
    \label{fig:teaser} 
\end{figure}

\section{Introduction}
A single image may depict a complex, dynamic scene involving many entities interacting with each other; human viewers are able to understand the gestalt of the high-level situation being shown as well as identifying its constituent inter-entity interactions. For example, an image of a football game may be recognized via both global scene cues (field, stadium setting, and crowd in the background) and interactions between entities in the image (players tackling, passing, running, etc.). 
Automatic understanding of such images shows promise for a variety of tasks, with applications to fields such as robotics, content retrieval, security systems, and assistive technology~\citep{stergiou2019analyzing, liu2023open, robinson2023robotic}. However, machine learning models currently lag far behind human performance on these tasks.
The current best-performing approaches are tailored to individual sub-tasks, commonly using unique architectures and being trained on different data. This includes tasks requiring overall understanding of the global semantics of images such as Situation Recognition (\sr)~\citep{yatskar2016situation}, and Human-Human Interaction (HHI) understanding~\citep{alper2023learning}, as well as those grounded in localized image regions such as Human-Object Interaction (HOI) detection~\citep{zhang2023pvic} and Grounded Situation Recognition (GSR)~\citep{pratt2020grounded}. By contrast to the fragmented ecosystem of different architectures for these tasks, there is a common thread of requiring an understanding of dynamic situations occurring in images, suggesting that they might benefit from a shared approach to high-level semantics and composition of images depicting dynamic activities and complex scenes.

In parallel to these works, recent breakthroughs in multimodal learning have used web-scale datasets of paired images and text to learn extremely diverse semantics of images~\citep{radford2021learning,rombach2022high}. In particular, vision representations trained jointly with language data have been shown to be powerful when used as-is for vision tasks, or when parsed with large language models (LLMs) for cross-modal understanding~\citep{zhai2022lit,alayrac2022flamingo,li2023blip,liu2024visual}. Underlying these works is the finding that language accompanying images encodes an understanding of their semantics and the situations which they depict, and that the use of these strong representations may obviate task-specific engineering.

\begin{figure*}[t]
    \center
    \jsubfig{\includegraphics[width=1\textwidth]{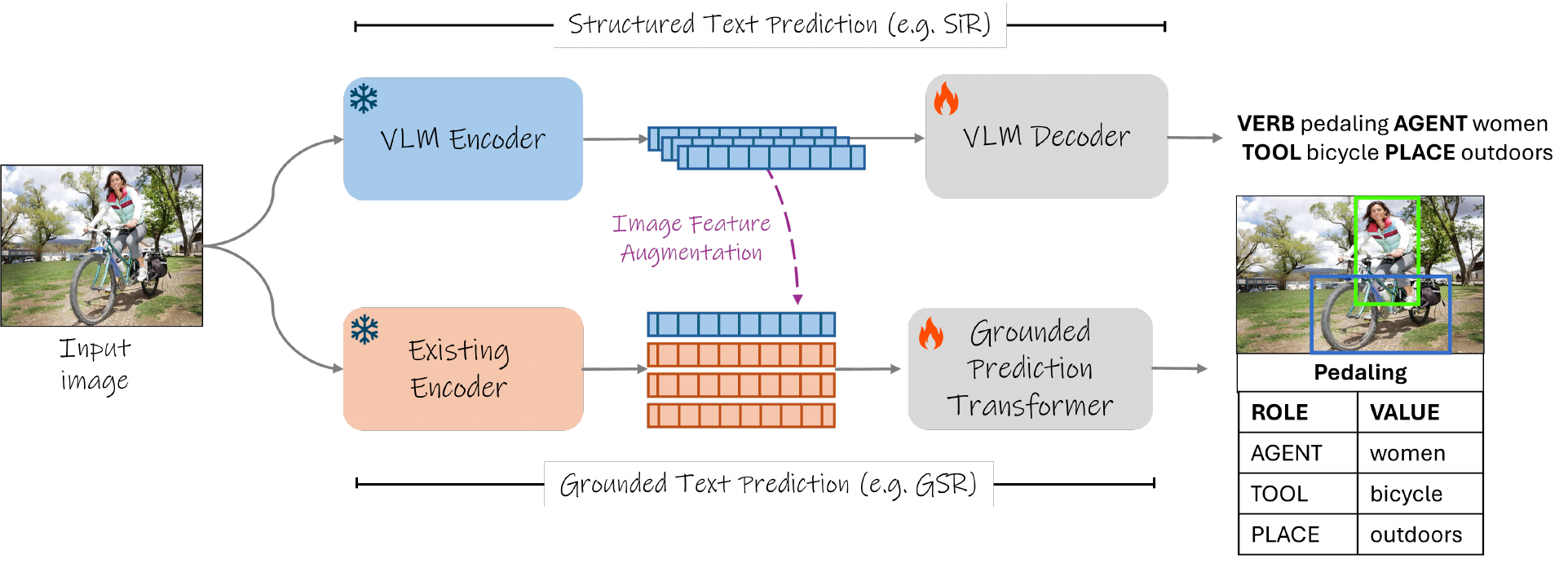}
    }   
    
    \caption{\textbf{Our framework.} We illustrate our framework for performing high-level (top) and grounded (bottom) tasks, exemplified by the tasks of \sr and GSR in the figure. For overall scene understanding tasks we add trainable weights to the VLM text decoder and fine-tune these using a standard token-wise language modeling objective to predict the desired labels as formatted text. For grounded prediction tasks we concatenate the V\&L embeddings to the existing vision backbone embeddings. The existing model is trained according to its initial formulation.}
    \label{fig:system}
    
\end{figure*}

In this work, we bridge the gap between these advancements and the existing state of dynamic scene understanding tasks, by adopting the use of pretrained vision-language (V\&L) representations as a powerful prior on the semantics of complex scenes in images.
By using frozen vision representations, we achieve SOTA across dynamic scene understanding tasks (\sr, HHI, HOI, GSR) with a framework requiring minimal task-specific engineering, as illustrated in Figure \ref{fig:system}.
For high-level (non-grounded) image understanding, we frame tasks as predicting and parsing structured text, using LLM-based decoding to extract semantic frames and descriptions of interactions in images. For fine-grained, grounded prediction, we augment the features extracted by existing models' vision backbones with these frozen representations for localized prediction of interactions. Across the board, this yields state-of-the-art results surpassing the best-performing existing models on standard benchmarks.
We also analyze the dynamic knowledge of these representations, showing that recent, more powerful V\&L representations encode dynamic scene dynamics which correlates with overall performance on the tasks under consideration.
Our findings show the promise of a unified approach towards related tasks concerning dynamic scenes in images by leveraging the knowledge of pretrained multimodal foundation models.

\section{Related Work}

\noindent
\textbf{Dynamic scene understanding in images}.
Due to the importance of understanding the semantics of images depicting complex, dynamic scenes, a number of works have focused on particular sub-tasks requiring the prediction of structured data from such images.  \sr refers to the prediction of an action verb and its semantic roles given a still image as input~\citep{yatskar2016situation}; GSR expands on this by incorporating bounding box predictions for each grounded entity~\citep{pratt2020grounded}. Subsequent works have greatly improved performance on these tasks using modern transformer architectures~\citep{wei2022rethinking, cho2022collaborative, roy2024clipsitu}. While these tasks consider a single overall situation to describe an image, this may not fully capture localized interactions between entities, which are of particular interest in applications such as robotics. As such, many works have focused on HOI detection in images, which requires both grounding (localizing humans and objects) and pairwise prediction of interactions between these entities~\citep{Tamura2021qpic, Iftekhar2022semantic, zhang2022upt,zhang2023pvic,yuan2023rlipv2}. %
In the particular case of interactions between humans, earlier works consider this as a categorical classification problem on the image level~\citep{yang2012recognizing,xiong2015recognize,antol2014zero}, while Alper \etal ~\citep{alper2023learning} propose to frame HHI understanding as an image captioning task with targets as unconstrained free text due to the non-local and context-dependent nature of HHI.
While all of these tasks consider images with dynamic contents, the particular methodologies used differ considerably, while our framework may be applied to all of these tasks.

We also note differences with other lines of work that also consider structured visual understanding. One line of work performs action recognition directly on video data~\citep{tang2012learning, tran2015learning, wang2021actionclip, zhu2020study, carreira2017quo}; however, like the works cited above, we consider the challenging case of a single still image. Another extensive line of work predicts scene graphs from images~\citep{chang2021comprehensive,zhu2022scene}; these consider a limited set of relation types (e.g. relative location) which mostly do not capture complex, dynamic interactions in images.
Finally, Chen et al.~\citep{chen-etal-2023-vistruct} propose a visual programming approach to situation recognition bearing some similarity to our structured image description paradigm; however, their approach involves the additional complexity of iterative program generation and execution without addressing the variety of tasks which our framework considers.

\medskip
\noindent
\textbf{Transfer learning from V\&L representations}.
Recent years have seen a shift from methods tailored to individual vision tasks, towards the use of transfer learning applied to pretrained foundation models~\citep{bommasani2021opportunities}. Web-scale vision-language training has shown to produce representations that excel at tasks such as zero-shot image classification, image captioning and VQA~\citep{jia2021scaling,radford2021learning,li2023blip}, and that may be processed with large language models (LLMs) to further reason over visual content~\citep{alayrac2022flamingo,liu2023llava, bai2023qwen, Huang2023Language, li2023Otter}.
Such representations have also been found to transfer effectively to tasks requiring grounded prediction within images, such as pixel-level segmentation~\citep{luddecke2022image,zhou2023zegclip,wang2022cris,lin2023clip}, referring expression grounding~\cite{peng2023Kosmos}, and grounded image generation~\cite{Sun2023Generative}.
Regarding dynamic scene understanding, prior works have leveraged CLIP for HOI detection~\citep{ning2023hoiclip,mao2024clip4hoi} and \sr~\citep{roy2024clipsitu}, showing the benefit of pretrained vision-language representations for these tasks. However, they consider isolated sub-tasks of dynamic scene understanding and still design task-specific architectures.
Our work advances this research direction by using SOTA multimodal representations and by unifying multiple situation understanding tasks, both those requiring high-level understanding (\sr, HHI) and those requiring localized, grounded predictions (HOI, GSR).

\section{Method}
We propose a framework for using frozen multimodally-pretrained vision representations to perform dynamic scene understanding tasks. This consists of two complementary methods for overall scene understanding and grounded prediction respectively; these are illustrated schematically in Figure \ref{fig:system}. Our method is agnostic to the vision representations used, but as shown in our ablations, the empirically best-performing setup uses localized embeddings extracted from a BLIP-2~\citep{li2023blip} model.

\subsection{\TextMethod}
\label{sec:si_des}

For tasks involving global, high-level understanding of the situation occurring in an image (\sr, HHI), we propose a dramatic simplification relative to prior works by predicting image attributes as structured text (Figure \ref{fig:system}, top). By outputting a single textual prediction per image and then parsing this structured text into the desired format, we are able to reformulate these existing tasks as image captioning and apply a standard image-to-text approach. We fine-tune an existing text decoder with a small number of trainable weights on the train set of each respective dataset, converted into the relevant structured text format, using a standard token-wise language modeling objective. Specifically, given an image representation $I$ and its corresponding structured text $T = (t_1, t_2, \ldots, t_n)$ composed of tokens $t_i$, we aim to maximize the log-likelihood:
$\mathcal{L}_{\text{LM}} = \sum_{i=1}^{n} \log P(t_i \mid I, t_1, t_2, \dots, t_{i-1}; \theta)$ where $P(t_i \mid I, t_1, t_2, \ldots, t_{i-1}; \theta)$ is the probability of the $i$-th token given the image representation and the prior tokens, and $\theta$ represents the parameters of the model. 

To convert between raw text and structured predictions, we use a task-specific parser component incorporating rule-based logic. For example, on the \sr{} task, an image is described by semantic frame data given by a verb $v$, its semantic roles $[r_1, r_2, . . . , r_m]$, each filled by an entity denoted by its noun value $N = [n_1, n_2, . . . , n_m]$ (or empty $\emptyset$). We parse this into a text description as \texttt{VERB} and verb $v$ in present continuous form (``-ing''; e.g. \texttt{VERB eating}), followed by concatenated pairs $(r_i, n_i)$ (e.g. \texttt{AGENT man}), where each role text $r_i$ is written in capital letters. This creates a single string (e.g. \texttt{VERB slicing AGENT person PLACE table TOOL knife}) which can be parsed unambiguously into and from semantic frame data. We parse semantic frames into such structured strings for supervising the finetuning of the text decoder. During inference, we can parse the structured text prediction (for instance, as shown in the top right corner of Figure \ref{fig:system}) into a semantic frame data (Figure \ref{fig:system}, bottom right).

\subsection{\GroundedMethod}
\label{sec:feat_concat}

For tasks involving spatially-grounded predictions (HOI and GSR), the output format must consist of bounding boxes and their attributes and relations, which is not easily parsed into a single, concise text-based description. However, existing architectures have been designed to output predictions of this format, typically using an existing backbone to extract grounded image features used as input to attention mechanisms of all subsequent stages of processing. In order to augment existing transformer-based models with knowledge from additional frozen V\&L representation, we propose to inject these through our \emph{\attentionmethod} mechanism, which concatenates them to the existing features extracted by the model's vision backbone. By contrast, replacing the model's backbone with the frozen VLM encoder would degrade fine-grained localization abilities (as large-scale V\&L representations are not expected to contain precise grounding information, as shown in our analysis on the GSR task on Section \ref{sec:ablation}, Table \ref{tab:misc3_results}). %
Intuitively, this provides additional dynamic situational knowledge to the existing model (Grounded Prediction Transformer in Figure \ref{fig:system}), while providing it with strictly more grounded knowledge for localized predictions. As seen in Figure \ref{fig:system}, this concatenation is performed by first projecting the new embeddings to the same size as the existing backbone's features, followed by feature-wise concatenation. Formally, given two sets of embeddings, \( E_{\text{backbone}} \) (extracted by the model's backbone) and \( E_{\text{V\&L}} \) (from the frozen V\&L representation), we first project \( E_{\text{V\&L}} \) into the dimensionality of \( E_{\text{backbone}} \) using a projection function, \( \pi \). The final concatenated tensor, \( F_{\text{concat}} \), is given by:

\[
F_{\text{concat}} = \text{concat}(E_{\text{backbone}}, \pi(E_{\text{V\&L}})) \in \mathbb{R}^{B \times K \times N}
\]
\noindent
where \( B \) is the batch size, \( K \) the number of spatial locations, and \( N \) the feature dimension. 

As the weight dimensions of attention mechanisms are invariant to the number of input features, this effectively expands the model without requiring additional weights beyond a linear projection layer. We also note that we do not add extra positional encodings, assuming that these are already present within the existing and newly added features.

\begin{figure*}[t]
\center
    \includegraphics[width=0.99\textwidth]{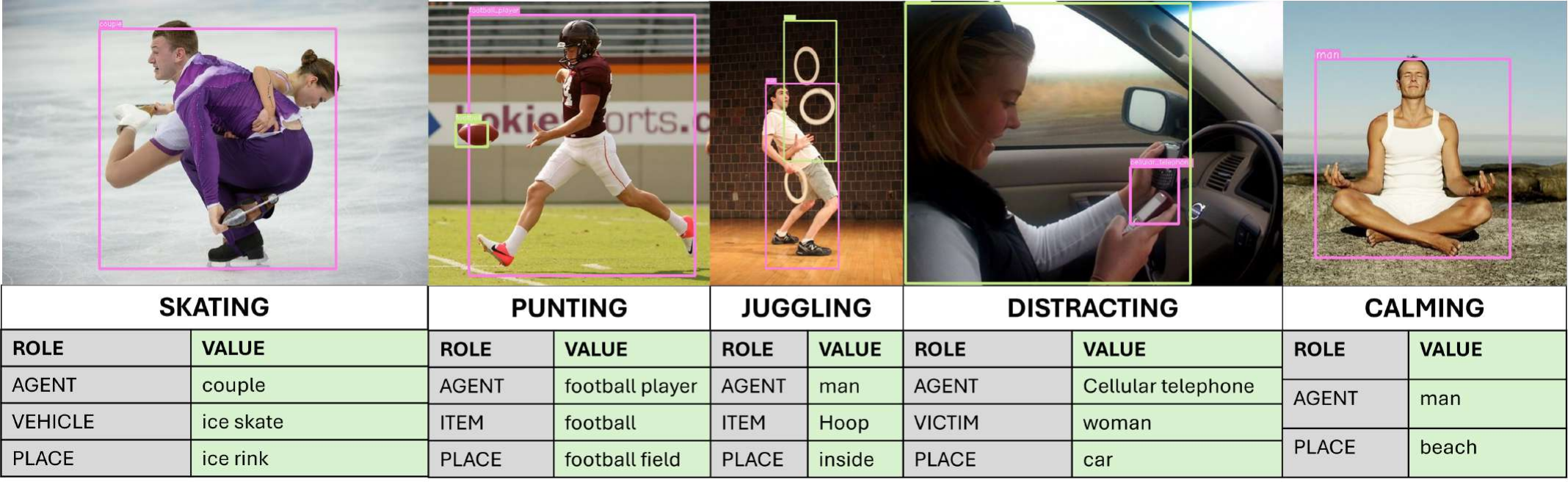}
    \caption{
    \textbf{Grounded Situation Recognition (GSR) qualitative results.} Results on the SWiG~\cite{pratt2020grounded} benchmark using our \attentionmethod method applied to CoFormer~\cite{cho2022collaborative}. The predicted main activity for image is indicated below it, while the corresponding predicted semantic roles (arguments) are displayed in the table, with nouns labeled according to their specific roles within the activity. Bounding boxes for the predicted AGENT are shown in \textcolor{VioletRed}{pink} and other predicted roles' boxes are shown in \textcolor{Green}{green}. As demonstrated, our method successfully predicts complex situations, including those involving non-human agents.
    }
    \label{fig:gsr_task_results} 
\end{figure*}

\begin{figure*}
  \jsubfig{\includegraphics[height=4.3cm]{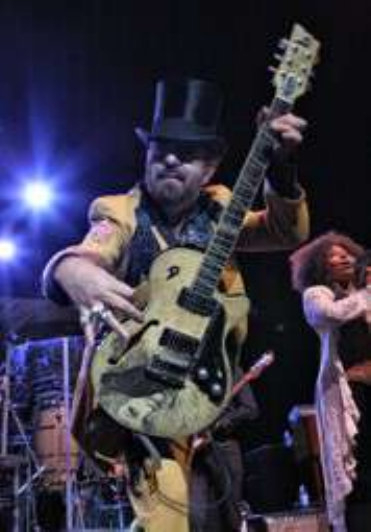}}
  {\vspace{-5pt}\begin{center} \small{(a) 
  
  [*] performing with [*]} \end{center}}
  \hfill
  \jsubfig{\includegraphics[height=4.3cm]{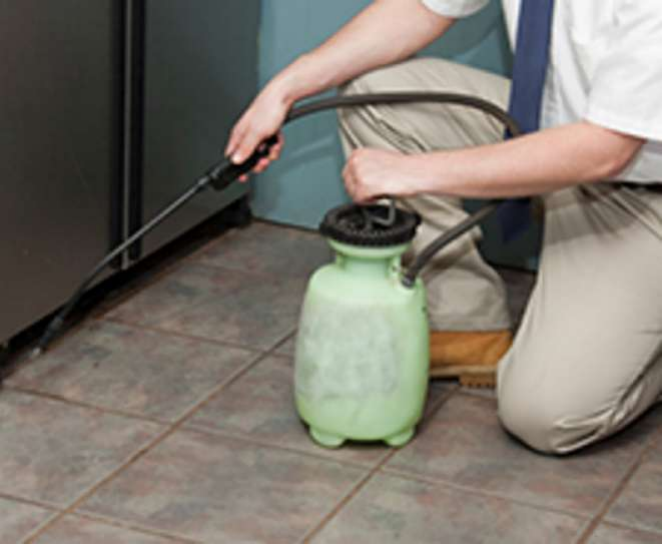}}
  {\vspace{-5pt}\begin{center} \small{(b)\\ VERB exterminating AGENT man INSTRUMENT nozzle PLACE floor} \end{center}}
  \hfill
  \jsubfig{\includegraphics[height=4.3cm]{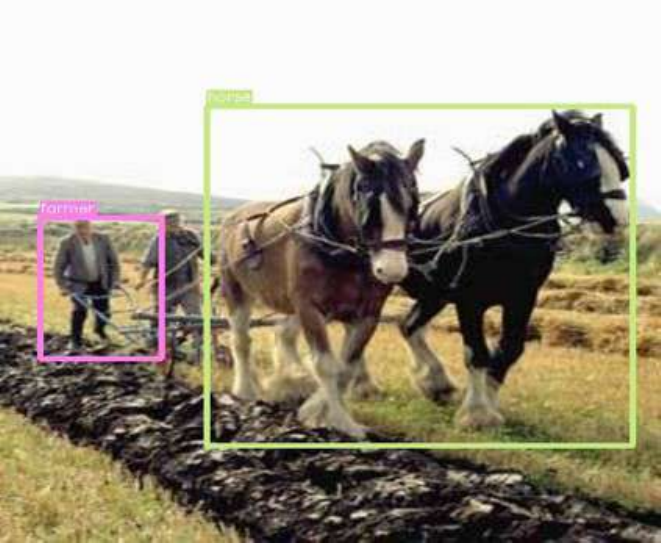}}
  {\vspace{-5pt}\begin{center} \small{(c)
\\ VERB plowing AGENT farmer INSTRUMENT horse PLACE field} \end{center}}
  \hfill
  \jsubfig{\includegraphics[height=4.3cm]{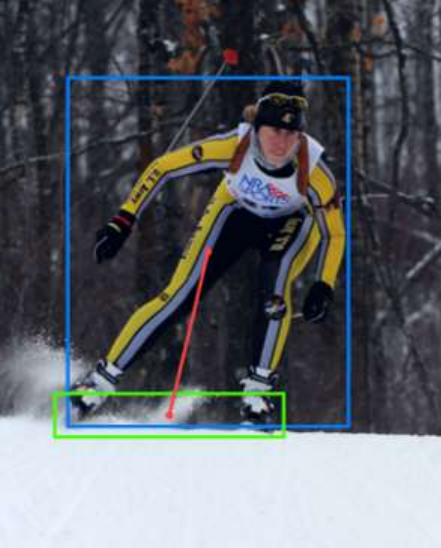}}
  {\vspace{-5pt}\begin{center} \small{(d)\\ride; stand on; wear} \end{center}}
  
  \caption{\textbf{Qualitative results.} 
  Results of our framework over several dynamic scene understanding tasks: \textbf{(a)} Human-Human Interaction (HHI), \textbf{(b)} Situation Recognition (SiR), \textbf{(c)} Grounded Situation Recognition (GSR), and \textbf{(d)} Human-Object Interaction Detection (HOI). For further details on these tasks, see Section \ref{sec:tasks}}   %
\label{fig:all_tasks_result}
\end{figure*}

\section{Tasks}
\label{sec:tasks}
\medskip
\noindent
We demonstrate our proposed framework on four different dynamic scene understanding tasks: human-object interaction detection
and grounded situation recognition, 
both of which are grounded prediction tasks, and human-human interaction recognition
and situation recognition,
both of which are text-only prediction tasks, treated as structured prediction tasks in our framework. 
Qualitative results of the four tasks are shown in Figure \ref{fig:all_tasks_result}.
We provide an overview of these tasks and their associated metrics below, with further details provided in the supplementary material. 

\medskip
\noindent
\textbf{Human-Object Interaction Detection (HOI).} The task of detecting human-object interactions involves localizing and classifying pairs of humans and objects interacting within an image~\cite{chao2018learning}. Interactions are typically represented as a triplet consisting of the object type, the specific interaction (action), and the corresponding bounding boxes for both the human and the associated object. The standard evaluation procedure uses a closed set of possible HOI classes combining fixed object and action categories~\citep{chao2018learning}.
Following prior work, we report average precision (AP) over joint bounding box and action predictions for three categories: a complete set of 600 HOI classes (full), a subset of 138 HOI classes with fewer than 10 training instances (rare), and a group of 462 HOI classes with 10 or more training instances (non-rare).

\medskip
\noindent
\textbf{Situation Recognition (\sr).} The goal of \sr is to generate a structured summary of an image that captures the primary activity and the entities involved in specific roles, forming a \emph{semantic frame} structure as defined in the field of linguistic semantics~\cite{yatskar2016situation}. In this formulation, the central activity being depicted corresponds to the chosen verb, whose \emph{arguments} are nouns labelled by their task in this action. For example, in Figure \ref{fig:sir_results}, the first image corresponds to a semantic frame containing fields such as \texttt{VERB = slicing} (central action depicted) and \texttt{TOOL = knife}. In this tabular structure, there exist a fixed set of possible keys defined in linguistic catalogues such as FrameNet~\citep{fillmore2003background}.
The metrics utilized for semantic role labeling are \emph{verb}, \emph{value} and \emph{value-all} ~\citep{yatskar2016situation}, which assess the accuracy of the verb and the noun predictions. For a verb with k roles, \emph{value} indicates whether the predicted noun matches at least one of the k roles. In contrast, \emph{value-all} evaluates whether all predicted nouns for all k roles are accurate. 

\medskip
\noindent
\textbf{Grounded Situation Recognition (GSR).} 
GSR extends the \sr task by additionally expecting a bounding box prediction for each nominal argument, i.e. requiring a predicted location for each participant in the action~\citep{pratt2020grounded}.
The situation localization metrics, grnd value and grnd value-all, evaluate the accuracy of bounding box predictions ~\citep{pratt2020grounded}, similar to the value and value-all metrics. A predicted bounding box is considered correct if it overlaps with the ground truth by 50\% or more. The metrics value, value-all, grnd value, and grnd value-all are assessed across three scenarios based on whether we are using the ground truth verb, the top-1 predicted verb, or the top-5 predicted verbs.

\medskip
\noindent
\textbf{Human-Human Interaction Recognition (HHI).} The task of understanding interactions between humans bears similarity to HOI detection, but has attracted separate attention and approaches due to the complex nature of HHI as depending on social context, their often non-local nature, and connection to human body pose~\citep{stergiou2019analyzing,alper2023learning}. While earlier work treated this as a categorical prediction task~\citep{ryoo2010ut,xiong2015recognize}, we follow the recent variant which predicts the most salient HHI in an image as free text~\citep{alper2023learning}. In this setting, model-generated text is evaluated relative to ground-truth text describing HHI using text generation metrics measuring qualities such as semantic similarity and factual groundedness.
We also follow their evaluation protocol, using BLEURT (BL)~\cite{sellam2020bleurt} for measuring textual similarity, NLI scores ($p_e$, $p_c$) for measuring factual groundedness, and verb embedding similarity (sim). %

\section{Experiments}
Below we present results for our method applied to the tasks from Section \ref{sec:tasks}, along with comparisons to existing models and ablations. We also perform an analysis over various V\&L representations, also comparing performance to a model that is pretrained in a unimodal fashion (without access to language). Additional implementation details, results and ablations are provided in the supplementary material.

\subsection{Experimental Design}

\noindent
\textbf{Modeling.}
For all tasks, we use vision representations from BLIP-2~\citep{li2023blip}, a VLM which has achieved SOTA performance on various multimodal tasks. This model is a vision-language encoder-decoder model consisting three components: (1) a ViT~\citep{dosovitskiy2020vit} vision encoder, (2) a LLM decoder, and (3) the Q-Former (QF) transformer bridging the vision-language modality gap. It has been pretrained on vision-language contrastive and generative tasks, after initializing the encoder and decoder from pretrained ViT and LLM models respectively. For all tasks, we use frozen embeddings output by components 1--2, consisting of $n=32$ unpooled embeddings of dimension $d=768$ corresponding to alternating image patches. For \textmethod, we use BLIP-2 with an OPT-2.7B~\citep{zhang2022opt} LLM decoder, while we discard this decoder for grounded predictions, as described below. We apply LoRA~\citep{hu2021lora} to the BLIP-2 LLM text decoder, adding a small number of trainable weights. For \groundedmethod, we apply \attentionmethod with these BLIP-2 embeddings to existing SOTA models: PViC~\citep{zhang2023pvic} (using Swin-L and H-DETR) and CoFormer~\citep{cho2022collaborative}, for HOI detection and GSR respectively.

\noindent
\textbf{Datasets and comparisons.} Our experiments cover the datasets HICO-DET~\citep{chao2018learning} (HOI), imSitu~\citep{yatskar2016situation} (\sr), SWiG~\citep{pratt2020grounded} (GSR), and Waldo and Wenda~\citep{alper2023learning} (HHI), using the standard metrics from each (see Section \ref{sec:tasks}). We compare to leading methods for each task, including the existing models PViC and CoFormer used as the base for \attentionmethod. Comparisons to ClipSitu XTF~\cite{roy2024clipsitu} use the variant with best performance on verb prediction.

\subsection{Evaluation}

\begin{table}[t]
    \centering
    \small
    \setlength{\tabcolsep}{4pt} %
    \renewcommand{\arraystretch}{0.9} %
    \begin{tabular}{lccc}
        \toprule
        Method & Verb & Value & Value-All \\
        \midrule   SituFormer~\cite{wei2022rethinking} & 44.20 & 35.24 & 21.86 \\
        CoFormer~\cite{cho2022collaborative} & 44.66 & 35.98 & 22.22 \\
        ClipSitu XTF~\cite{roy2024clipsitu} & 58.19 & 47.23 & 29.73 \\
        \midrule
        Ours & \textbf{58.88} & \textbf{51.10} & \textbf{31.56} \\
        \bottomrule
    \end{tabular}
    \caption{\textbf{\sr Evaluation.} Models evaluated on the imSitu dataset~\citep{yatskar2016situation} (test set), evaluated for predictions of verbs and their semantic arguments.
    }
    \label{tab:sir_results_main}
\end{table}

\begin{table}[t]
    \centering
    \small
    \setlength{\tabcolsep}{4pt} %
    \renewcommand{\arraystretch}{0.9} %
    \begin{tabular}{lcccc}
        \toprule
        Method & BL $\uparrow$ & $p_e \uparrow$ & $p_c \downarrow$ & sim $\uparrow$  \\ 
        \midrule
        EncDec~\citep{alper2023learning} & 0.38 & 0.30 & 0.37 & 0.41  \\ 
        CLIPCap~\citep{mokady2021clipcap} & 0.42 & 0.40 & 0.32 & 0.46 \\ 
        \midrule
        Ours & \textbf{0.46} & \textbf{0.45}& \textbf{0.30} & \textbf{0.53} \\ 
        \bottomrule
    \end{tabular}
\caption{\textbf{HHI quantitative results.} Results on the Waldo and Wenda benchmark ~\citep{alper2023learning}. Following prior work~\citep{alper2023learning}, we report performance over
BLEURT (BL) and NLI scores ($p_e$, $p_c$) and verb embedding similarity (sim). 
} %
\label{tab:hhi_results}
\end{table}

\begin{figure*}
    \centering
    \includegraphics[height=5.5cm]{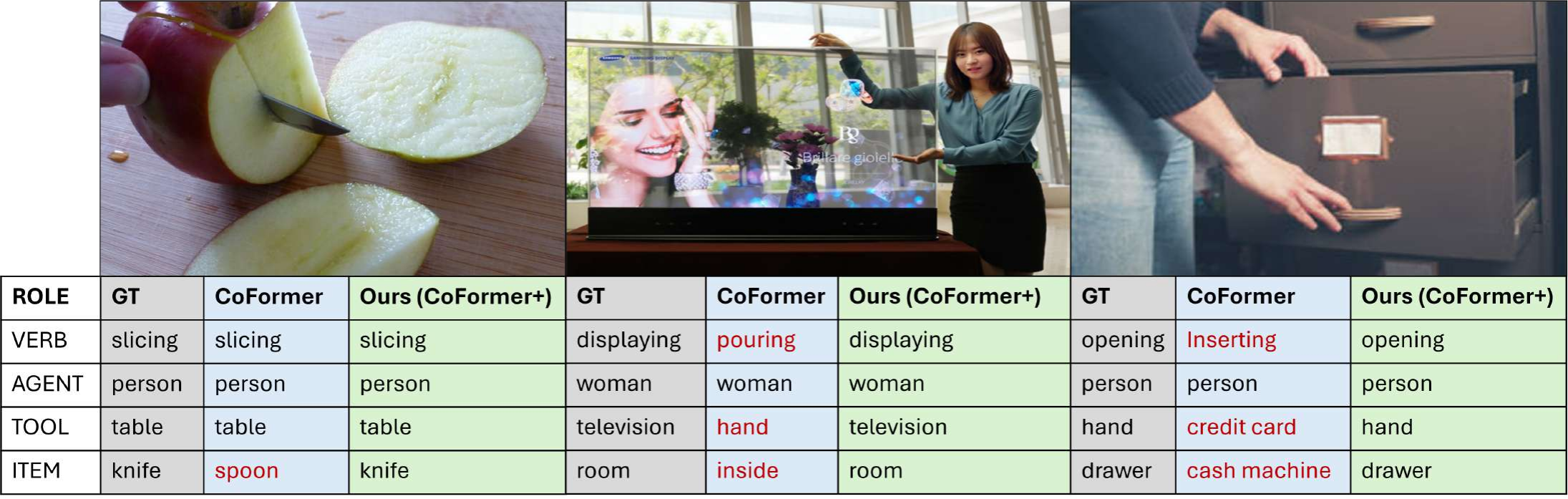}
    \caption{\textbf{Situation Recognition (SiR) qualitative comparison.}
Comparison of results between CoFormer~\cite{cho2022collaborative} and our method (CoFormer+) on the imSitu~\cite{yatskar2016situation,pratt2020grounded} test set. We apply our our proposed \attentionmethod mechanism to a CoFormer backbone. Incorrect predictions are shown in \textcolor{BrickRed}{red}.}
    \label{fig:sir_results}
\end{figure*}

\begin{figure*}
  \jsubfig{\includegraphics[height=3.5cm]{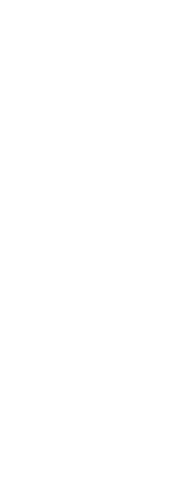}}{\begin{flushleft}\vspace{-5.5pt} \small{\footnotesize \textbf{Ours} \newline \vspace{-5pt} \newline \footnotesize \textbf{CoFormer} \newline \vspace{-5pt} \newline \footnotesize \textbf{CLIPCap} \newline \vspace{-5pt} \newline \footnotesize \textbf{GT}}
  \end{flushleft}} 
  \hfill
  \jsubfig{\includegraphics[height=2.9cm]{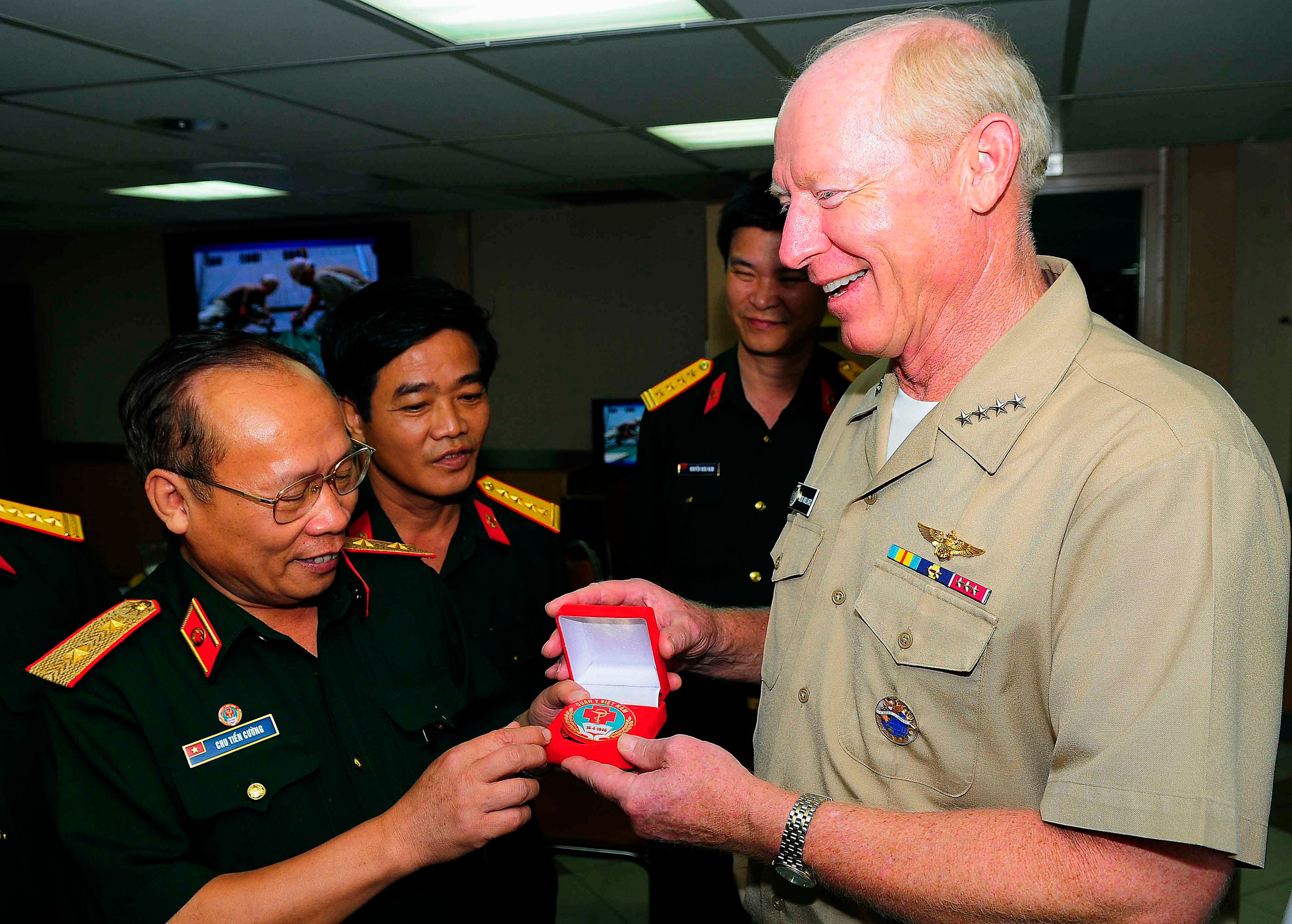}}{\vspace{-5pt}\begin{flushleft} \small{[*] giving a medal to [*]\newline \vspace{-5pt} \newline \textcolor{red}{admiring} \newline \vspace{-5pt} \newline [*] giving [*] a present \newline \vspace{-6pt} \newline [*] receiving a command coin
from [*]} \end{flushleft}}
  \hfill
  \jsubfig{\includegraphics[height=2.9cm]{figures/results/COCO_val2014_000000384468.pdf}}{\vspace{-5pt}\begin{flushleft} \small{[*] riding an elephant with [*] \newline \vspace{-5pt} \newline \textcolor{red}{stroking} \newline \vspace{-5pt} \newline [*] being carried by [*] \newline \vspace{-6pt} \newline [*] riding an elephant with [*]} \end{flushleft}}
  \hfill
  \jsubfig{\includegraphics[height=2.9cm]{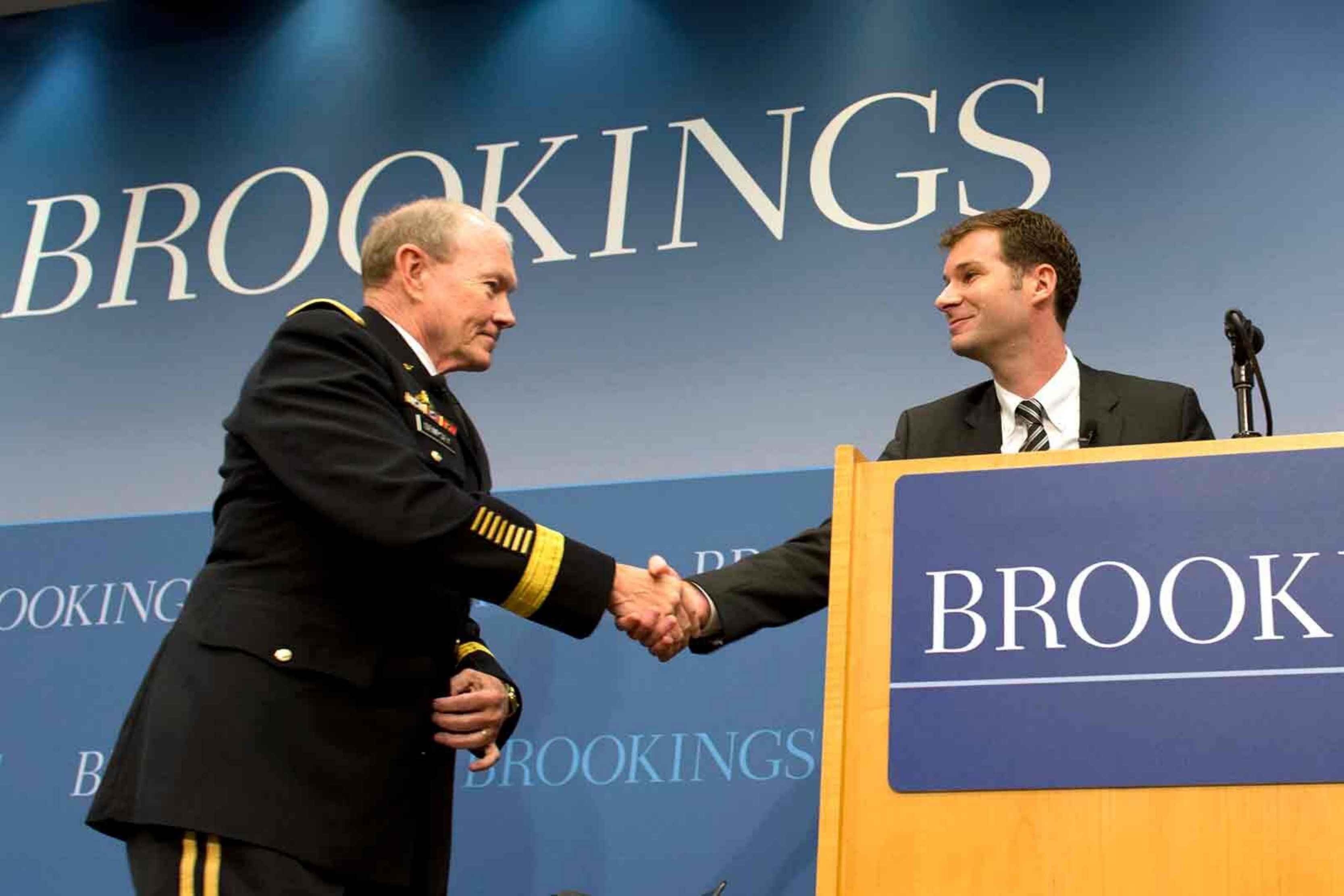}}{\vspace{-5pt}\begin{flushleft} \small{[*] shaking hands with [*] \newline \vspace{-5pt} \newline  shaking \newline \vspace{-5pt} \newline \textcolor{red}{[*] speaking with [*]} \newline \vspace{-6pt} \newline [*] shaking hands with [*]} \end{flushleft}}
  \hfill
  \jsubfig{\includegraphics[height=2.9cm]{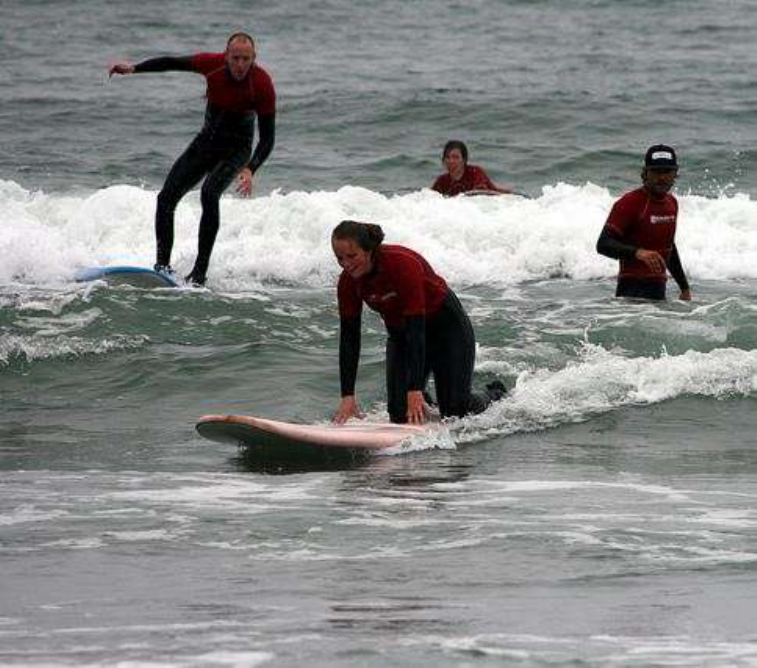}}{\vspace{-5pt}\begin{flushleft} \small{[*] surfing with [*] \newline \vspace{-5pt} \newline surfing  \newline \vspace{-5pt} \newline \textcolor{red}{[*] swimming with [*]} \newline \vspace{-5pt} \newline [*] surfing with [*]} \end{flushleft}}
  
  \caption{\textbf{HHI qualitative comparison.} Prior work~\citep{alper2023learning} vs. our results on the Waldo and Wenda test set~\citep{alper2023learning}, along with the ground-truth labels for reference. Incorrect predictions are denoted in \textcolor{red}{red}, while semantically-close predictions are not marked. }\label{fig:hhi_results2}
\end{figure*}

\noindent \textbf{\TextMethod Tasks}. Results on the \sr and HHI tasks are shown in Tables \ref{tab:sir_results_main}--\ref{tab:hhi_results}, comparing our method to SOTA approaches. For \sr, our method outperforms leading methods at predicting verbs and semantic frame participants in images, while maintaining conceptual simplicity. On the HHI task, we outperform prior captioning-based
approaches on semantic adequacy metrics (BLEURT, verb similarity) and factuality metrics ($p_e$, $p_c$). We further analyze qualitatively in the supplementary material. We also provide qualitative results in Figures \ref{fig:sir_results}--\ref{fig:hhi_results2}, illustrating the effectiveness of our approach.

\begin{table}[t]
    \centering
    \small
    \setlength{\tabcolsep}{4pt} %
    \renewcommand{\arraystretch}{0.9} %
    \resizebox{\columnwidth}{!}{%
    \begin{tabular}{lcccccc}
        \toprule
        \multicolumn{1}{c}{Method} & \multicolumn{2}{c}{Top-1 Verb} & 
        \multicolumn{2}{c}{Top-5 Verbs} & \multicolumn{2}{c}{GT Verb} \\
        \cmidrule(lr){2-3} \cmidrule(lr){4-5} \cmidrule(lr){6-7}
         & value & val-all & value & val-all & value & val-all \\
        \midrule
        SituFormer~\cite{wei2022rethinking} & 29.22 & 13.41 & 46.00 & 20.10 & 61.89 & 24.89 \\
        CoFormer~\cite{cho2022collaborative} & 29.05 & 12.21 & 46.25 & 18.37 & 60.11 & 22.12 \\
        ClipSitu XTF~\cite{roy2024clipsitu} & 40.01 & 15.03 & 49.78 & \textbf{25.22} & 53.36 & \textbf{33.20} \\
        \midrule
        Ours (CoFormer+) & \textbf{41.32} & \textbf{18.16} & \textbf{57.02} & 24.72& \textbf{65.11} & 27.28 \\
        \bottomrule
    \end{tabular}%
    }
    \caption{\textbf{GSR evaluation.} Above, we report performance on the test set of the SWiG dataset~\citep{pratt2020grounded}, with metrics measuring accurate verb and grounded argument prediction.
    }
    \label{tab:sir_results_grnd}
\end{table}

\begin{table}[t]
\centering
\setlength{\tabcolsep}{4pt}  %
\def\arraystretch{0.9}       %
\small                       %
\begin{tabular}{lccc}
    \toprule
    Method & Full & Rare & Non-rare \\ 
    \midrule
    
    RLIPv2~\cite{yuan2023rlipv2} &  45.09 & 43.23 & 45.64 \\   PViC w/ H-DETR & 44.32 & 44.61 & 44.24 \\ 
    \midrule

    Ours (PViC+) & \textbf{46.49} & \textbf{47.43} & \textbf{46.21} \\ 
    \bottomrule
\end{tabular}
\caption{\textbf{HOI quantitative results.} Comparison of detection performance on the HICO-DET test set, with metrics following Zhang et al.~\citep{zhang2023pvic}. All methods above use a Swin-L vision backbone; additional (underperforming) methods and backbones are provided in the supplementary material.
}
\label{tab:hoi_results}
\end{table}

\begin{table}[t]
    \centering
    \small
    \setlength{\tabcolsep}{4pt} %
    \renewcommand{\arraystretch}{0.9} %
    \resizebox{\columnwidth}{!}{%
    \begin{tabular}{lccccccc}
        \toprule
        \multirow{2}{*}{Backbone}  & \multirow{2}{*}{\shortstack{Attention Feature\\Augmentation}} &\multicolumn{2}{c}{Top-1} & 
        \multicolumn{2}{c}{Top-5} & \multicolumn{2}{c}{GT Verb} \\
        \cmidrule(lr){3-4} \cmidrule(lr){5-6} \cmidrule(lr){7-8}
         & & value & value-all & value & value-all & value & value-all \\
        \midrule
        R50~\cite{he2016deep} & x & 29.05 & 12.21 & 46.25 & 18.37 & 60.11 & 22.12 \\
        Blip2~\cite{li2023blip} & x & 27.27 & 6.44 & 38.21 & 8.62 & 43.5 & 9.5 \\
        R50 (Ours) & \checkmark & \textbf{41.96} & \textbf{19.62} & \textbf{59.04} & \textbf{26.56} & \textbf{67.06} & \textbf{29.15} \\
        \bottomrule
    \end{tabular}%
    }
    \caption{\textbf{GSR ablation results.} We evaluated GSR performance on Top-1 Predicted Verb, Top-5 Predicted Verb, and GT Verb with different backbones, with and without using our Attention Feature Augmentation method, showing that only replacing the backbone with a frozen VLM encoder degraded fine-grained localization.}
\label{tab:misc3_results}
\end{table}

\begin{figure*}
  \jsubfig{\includegraphics[height=1.6cm]{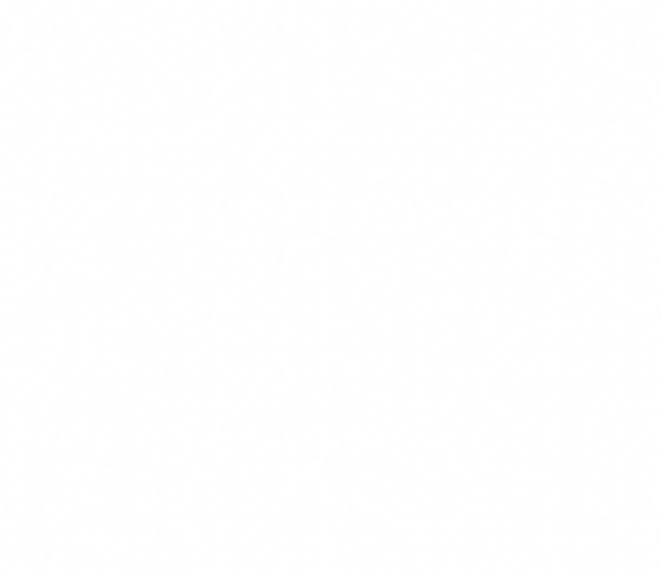}}{\begin{flushleft}\vspace{-6pt} \small{ Ours \footnotesize  (PViC+) \newline \vspace{-4pt}  \newline PViC \newline \vspace{-17pt} \newline \newline GT}
  \end{flushleft}} 
  \hfill
  \jsubfig{\includegraphics[height=3.6cm]{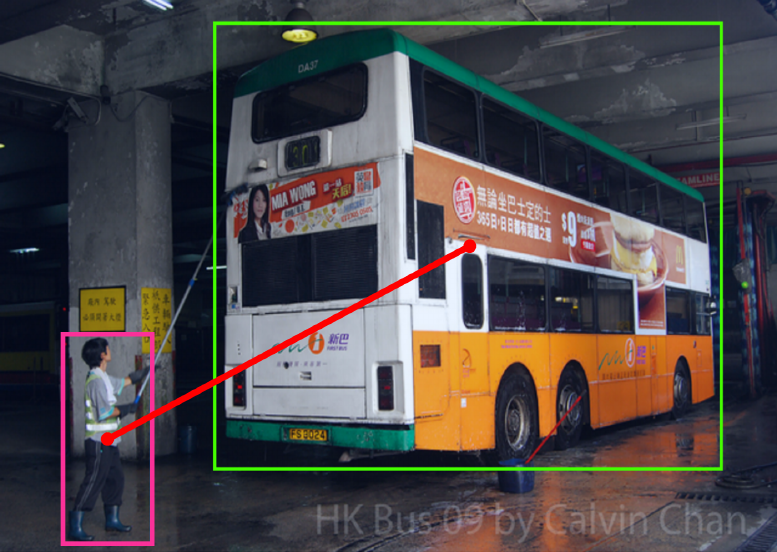}}{\vspace{-5pt}\begin{flushleft} \small \textcolor{red}{wash}
\\ \vspace{5pt}\textcolor{red}{-}  \\ \vspace{5pt} \textcolor{red}{washing a bus}  \\ %
  \end{flushleft}}
          \hfill
  \jsubfig{\includegraphics[height=3.6cm]{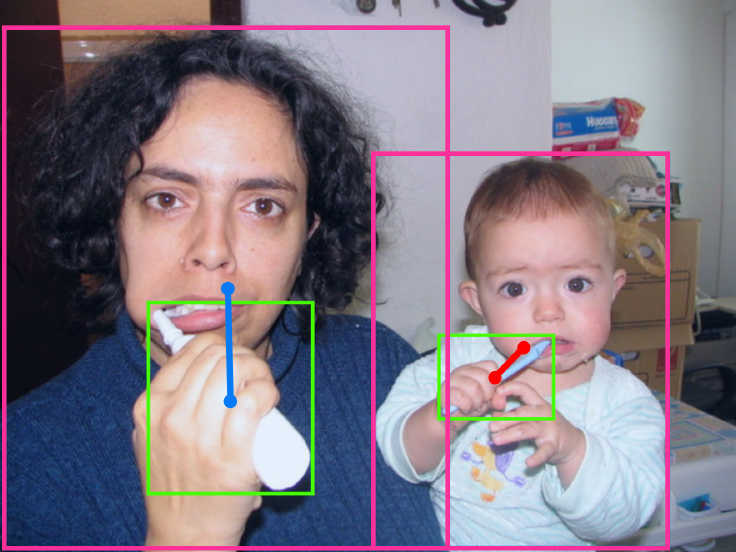}}{\vspace{-5pt}\begin{flushleft} \small   \textcolor{blue}{ brush with, hold;} \textcolor{red}{brush with, hold} \\ \vspace{5pt}\textcolor{blue}{hold; }\textcolor{red}{ hold}  \\ \vspace{5pt}\textcolor{blue}{brushing with a toothbrush, holding
  \\a toothbrush;}  \textcolor{red}{brushing with a toothbrush, holding a toothbrush} \end{flushleft}}
          \hfill
  \jsubfig{\includegraphics[height=3.6cm]{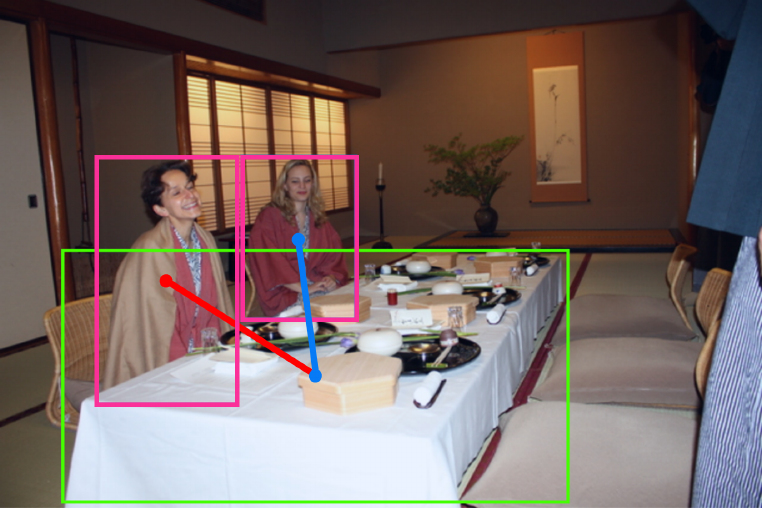}}{\vspace{-5pt}\begin{flushleft} \small{ \textcolor{blue}{sit at, eat at;} \textcolor{red}{sit at, eat at} \\ \vspace{5pt}  \textcolor{blue}{sit at;} \textcolor{red}{sit at}  \\\vspace{5pt}\textcolor{blue}{sitting at a table, eating at a table;} \textcolor{red}{sitting at a table, eating at a table} } \end{flushleft}}
          \hfill
  \caption{\textbf{HOI qualitative Comparison.} Example predictions of our \attentionmethod   (first row) and PViC~\cite{zhang2023pvic} (second row) on the HICO-DET~\cite{chao2015hico} test set.
    Predicted human and object bounding boxes are shown in \textcolor{Magenta}{pink} and \textcolor{ForestGreen}{green} respectively, and predicted interactions as a \textcolor{red}{red} and \textcolor{blue}{blue} lines, based on the description's interaction verbs colors.
    }
    \label{fig:hoi_results}
    \end{figure*}

\medskip
\noindent \textbf{\GroundedMethod Tasks} Results for GSR and HOI are reported in Tables \ref{tab:sir_results_grnd}--\ref{tab:hoi_results}. Our approach (\attentionmethod applied to CoFormer and PViC respectively) outperforms previous SOTA across almost all metrics, indicating consistent improvements at identifying and localizing actions and interactions in images, as well as their (both human and non-human) participants. In particular, our method provides a boost in performance to the existing SOTA models CoFormer and PViC relative to their use without our approach. Moreover, for HOI our method shows SOTA performance across full, rare, and non-rare categories, indicating strong performance on the long tail of possible interaction types. We also note that our comparison in Table \ref{tab:hoi_results} is over models using a strong Swin-L transformer backbone (with ResNet-50 results in the supplementary material). Qualitative results are provided in Figures \ref{fig:gsr_task_results} and \ref{fig:hoi_results}, further illustrating this strong performance.

\begin{figure}[t]
\center
    \includegraphics[width=0.5\textwidth]{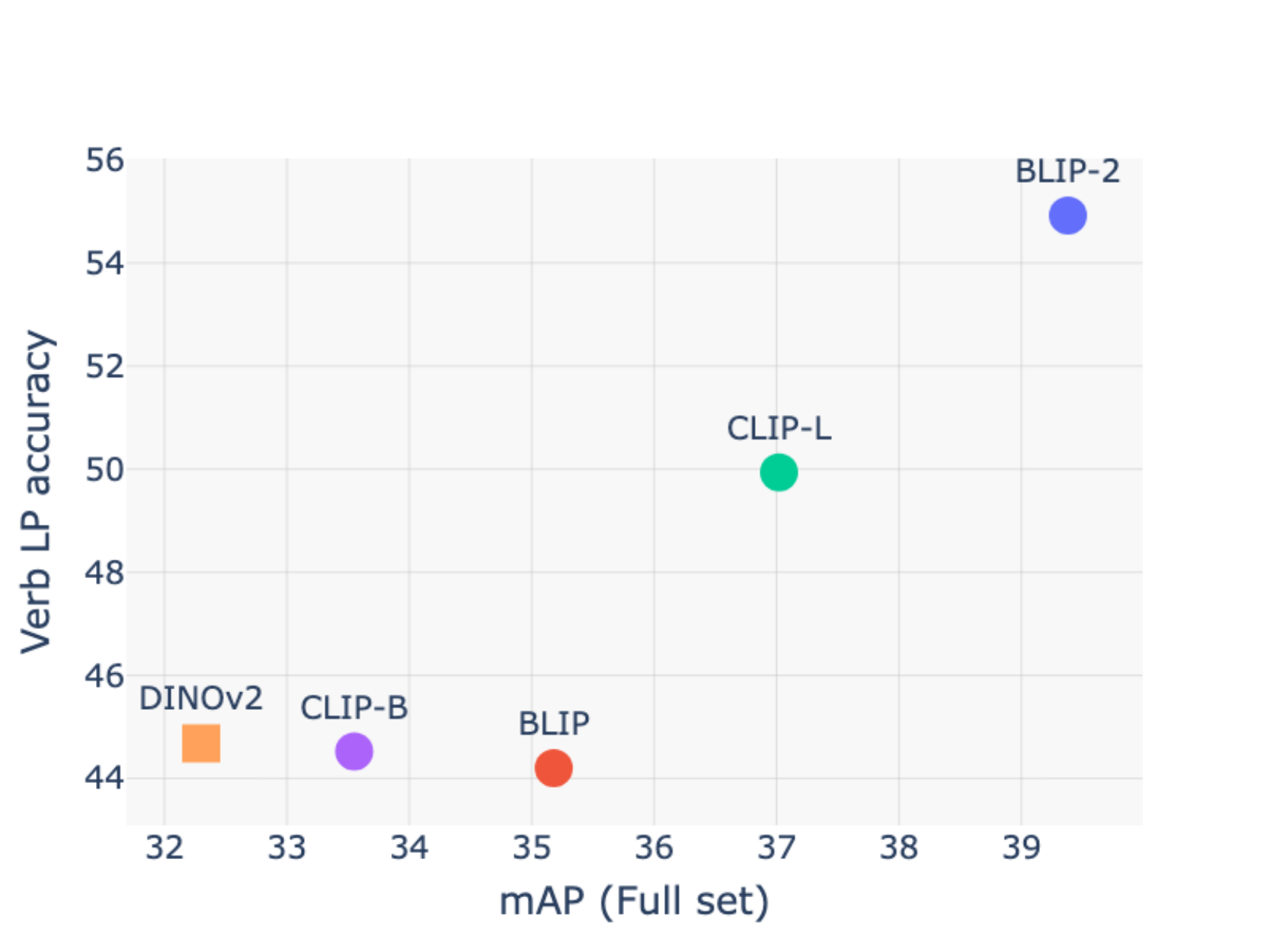}
    \caption{
    \textbf{Dynamic Knowledge in V\&L Representations.}
    We compare linear probing results for verb prediction (LP, y-axis) and overall performance on the HOI task (x-axis) across a range of vision embeddings. We find that embeddings linearly encoding richer dynamic knowledge, as measured by the LP results, perform strongly on dynamic tasks such as HOI prediction, with the multimodal (language-supervised) embeddings of BLIP-2 achieving the best results across the board. 
    \footnotesize{Square and circular markers denote unimodal (vision-only) and V\&L-pretrained models respectively.}}
   
    \label{fig:scatter} 
\end{figure}

\medskip

\subsection{Analysis of V\&L Representations}

In addition to the experiments detailed above, we perform an analysis comparing various modern V\&L representations to better understand the dynamic knowledge contained within these embeddings and its effect on dynamic scene understanding tests. As our framework is agnostic to the backbone used for calculating embeddings, we apply our method to an array of vision embeddings. As the common aspect of our array of tasks is the presence of dynamic action in images, we hypothesize that the best-performing embeddings for our framework are those that best encode dynamics in general. As a proxy for this, we apply linear probing (LP) to predict verbs corresponding to images, testing whether such dynamic concepts are linearly separable in the embedding space. We then compare performance to our dynamic tasks to see if this correlates with downstream dynamic scene understanding.

Our tests span V\&L models of different sizes, including our CLIP~\cite{radford2021learning}, BLIP~\cite{li2022blip}, and BLIP-2~\cite{li2023blip} representations. We also evaluate DINOv2~\cite{oquab2023dinov2} representations, which are pretrained in a unimodal fashion (using images alone). Here we present results for unpooled (localized) embeddings and for the HOI task. For verb prediction, we evaluate using the imSitu~\cite{yatskar2016situation} benchmark. Results are shown in Figure \ref{fig:scatter}, illustrating the that superior performance of large V\&L models (CLIP-L, and particularly BLIP-2) correlates with our LP-based measure of dynamicness. Our results justify the use of BLIP-2 embeddings as they show the strongest encoding of image dynamics and perform most strongly accross tasks.

\subsection{Ablation Studies} \label{sec:ablation}
To justify our Attention Feature Augmentation approach, we compare using the new VLM encoder directly as the frozen backbone of existing models for grounded prediction. In particular, for GSR we compare  CoFormer~\cite{cho2022collaborative} with BLIP-2 backbone used directly, rather than concatenating these features to the existing ResNet-50 backbone. As seen in Table \ref{tab:misc3_results}, this far underperforms our approach, suggesting that fine-grained localization knowledge is less accessible in this setting since it was not encouraged by BLIP-2's pretraining objective. 

\section{Discussion and Conclusions}

To automatically understand images depicting complex, dynamic scenes, we have proposed the use of powerful vision-language representations to provide a generic approach to a number of related tasks, namely \sr, HHI, GSR and HOI. Our results have shown that such representations, frozen and used as-is, may provide SOTA performance on both high-level understanding and fine-grained grounded predictions, implemented respectively as structured text prediction and via our \attentionmethod method. Additionally, our analysis of these representations shows that recent, more powerful V\&L representations better encode dynamic scene semantics, correlating with performance on these tasks. Our results suggest that vision-language pretraining is an important tool for tackling tasks requiring such scene understanding, and we foresee its application to additional related tasks such as scene graph generation and prediction. 

Despite achieving state-of-the-art results across multiple tasks, various limitations should be considered. First, the size and computational requirements of our model make it less suitable for applications that require real-time online prediction, such as robotics and interactive systems. Although we leverage BLIP-2 embeddings effectively, our research leaves open an investigation of why it surpasses other models at dynamic scene understanding, and how VLM pretraining could be enhanced to improve such capabilities. Furthermore, grounded prediction tasks rely on existing backbone models rather than achieving an end-to-end unified solution. While our method is effective in multiple tasks, it also requires fine-tuning to fit each task or dataset.

Looking ahead, we foresee future work investigating the optimal pretraining method for VLMs to encourage such dynamic knowledge, and how to incorporate more precise grounding into their representations to obviate the need for existing backbones in grounded prediction tasks. 

\medskip
\noindent
\textbf{Acknowledgements} 
 We thank Moran Yanuka, Chantal Shaib and Yuval Alaluf for providing
helpful feedback.

{\small
\bibliographystyle{ieeenat_fullname}
\bibliography{egbib}
}

\clearpage
\begin{appendix}
    
\section{Additional Evaluations and Comparisons}
\subsection{Qualitative results}
Interactive visualization of results on all tasks considered in our work (HOI, HHI, GSR and \sr) is available via our project page: \href{https://tau-vailab.github.io/Dynamic-Scene-Understanding/}{https://tau-vailab.github.io/Dynamic-Scene-Understanding/}. There, we present visualizations and comparisons to existing models over a random subset of each tasks's test set.
\subsection{Quantitative Evaluations}  
We provide additional qualitative results for the HOI task in Table \ref{tab:hoi_results_app}, including comparisons to models using a ResNet-50 backbone as well as those using a Swin-L transformer backbone. We include results for our method applied to PViC~\cite{zhang2023pvic} with both backbones, seeing that while models using a transformer (Swin-L) backbone generally show stronger performance, our method affords an additional boost in performance in either case. Table~\ref{tab:sir_results_app} provides the complete results for \sr on the imSitu benchmark~\cite{yatskar2016situation}, including validation and test sets. Similarly, Table~\ref{tab:gsr_results_app} shows the full results of our method on the SWiG benchmark~\citep{pratt2020grounded}, including validation and test sets and additional metrics. These additional results show our consistently strong performance, consistent with the results in our main paper. We also present more detailed breakdowns of our results on the HHI, \sr, and GSR task benchmarks in Tables \ref{tab:gsr_results_app}--\ref{tab:hhi_results_aa}. In particular, Table \ref{tab:hhi_results_aa} shows full HHI results on the Waldo and Wenda benchmark~\cite{alper2023learning} split by data source -- Who’s Waldo (WW), Conceptual Captions (CC), and COCO Captions.

\section{Additional Details}

\subsection{Training Details}
We proceed to describe training procedures for each task using our method. Unless otherwise stated, all tasks use BLIP-2 as the multimodal backbone. We apply LoRA to its decoder (while keeping the visual encoder frozen) for \textmethod tasks; for \groundedmethod tasks, we insert embeddings from its (frozen) encoder using \attentionmethod and train the existing model (CoFormer or PViC) with its initial formulation. All training is conducted on a single NVIDIA RTX A5000 GPU.
\begin{table}[t]
\centering
\setlength{\tabcolsep}{4pt}  %
\def\arraystretch{0.9}       %
\small                       %
\begin{tabular}{lccc}
    \toprule
    Method & Full & Rare & Non-rare \\ 
    \midrule
    RLIP$^\text{R}$~\cite{Yuan2022RLIP} & 32.84 & 26.85 & 34.63\\GEN-VLKT$^\text{R}$~\cite{liao2022gen} & 33.75 & 29.25 & 35.10 \\
    RLIPv2$^\text{R}$*~\cite{yuan2023rlipv2} & 27.01 & 35.21 & 33.32 \\
    RLIPv2$^\text{S}$~\cite{yuan2023rlipv2} &  {45.09} & 43.23 & {45.64} \\
    PViC$^\text{R}$ w/ DETR & 34.69 & 32.14 & 35.45 \\ 
    PViC$^\text{S}$ w/ H-DETR & 44.32 & {44.61} & 44.24 \\ 
    \midrule
    Ours (PViC$^\text{R}$+) & 39.38 & 39.53 & 39.34 \\
    Ours (PViC$^\text{S}$+) & \textbf{46.49} & \textbf{47.43} & \textbf{46.21} \\ 
    \bottomrule
\end{tabular}
\caption{\textbf{HOI quantitative results.} Comparison of detection performance on the HICO-DET test set, with metrics following Zhang et al.~\citep{zhang2023pvic}.
\footnotesize* denotes metrics calculated with Extra Relations. $^\text{R}$ denotes models using a ResNet-50 backbone, and $^\text{S}$ denotes those using a Swin-L backbone.}
\label{tab:hoi_results_app}
\end{table}

\begin{table}[htbp]
    \centering
    \begin{tabular}{|c|l|c|c|c|}
        \hline
        Set & Method & verb & value & value-all \\
        \hline
        \multirow{4}{*}{dev} 
        & SituFormer~\citep{wei2022rethinking} & 44.32 & 35.35 & 22.10 \\
        & CoFormer ~\citep{cho2022collaborative} & 44.41 & 35.87 & 22.47 \\
        & ClipSitu XTF$^*$~\citep{roy2024clipsitu} & {58.19} & {47.23} & {29.73} \\
        \rowcolor{gray!20} & Ours & \textbf{58.83} & \textbf{52.13} & \textbf{31.67} \\
        \hline
        \multirow{4}{*}{test} 
        & SituFormer~\citep{wei2022rethinking} & 44.20 & 35.24 & 21.86 \\
        & CoFormer ~\citep{cho2022collaborative} & 44.66 & 35.98 & 22.22 \\
        & ClipSitu XTF$^*$~\citep{roy2024clipsitu} & {58.19} & {47.23} & {29.73} \\
        \rowcolor{gray!20} & Ours & \textbf{58.88} & \textbf{51.10} & \textbf{31.56} \\
        \hline
    \end{tabular}
    \caption{Full evaluation of the \sr on both validation (dev) and test set on the imSitu~\cite{yatskar2016situation} dataset.{\footnotesize $^*$ClipSitu XTF metrics are for the best-performing method presented for verb prediction.}}
    \label{tab:sir_results_app}
\end{table}

\begin{table*}[htbp]
    \centering
    \resizebox{\textwidth}{!}{%
    \begin{tabular}{|c|l|c|c|c|c|c|c|}
        \hline
        \multirow{2}{*}{Set} & \multirow{2}{*}{Method} & \multicolumn{2}{c|}{Top-1 Predicted Verb} & \multicolumn{2}{c|}{Top-5 Predicted Verbs} & \multicolumn{2}{c|}{Ground-Truth Verb} \\
        \cline{3-8}
         &  & grnd value & grnd value-all & grnd value & grnd value-all & grnd value & grnd value-all \\
        \hline
        \multirow{4}{*}{dev} 
        & SituFormer~\citep{wei2022rethinking} & 29.17 & 13.33 & 45.78 & 19.77 & {61.82} & 24.65 \\
        & CoFormer ~\citep{cho2022collaborative} & 29.37 & 12.94 & 46.70 & 19.06 & 61.15 & 23.09 \\
        & ClipSitu XTF$^*$~\citep{roy2024clipsitu} & \textbf{41.30} & {13.92} & {49.23} & {23.45} & 55.36 & \textbf{32.36} \\
        \rowcolor{gray!20} & Ours & 41.03 & \textbf{18.83} & \textbf{57.87} & \textbf{25.48} & \textbf{65.98} & {28.04}  \\
        \hline
        \multirow{4}{*}{test} 
        & SituFormer~\citep{wei2022rethinking} & 29.22 & 13.41 & 46.00 & 20.10 & 61.89 & 24.89 \\
        & CoFormer ~\citep{cho2022collaborative} & 29.05 & 12.21 & 46.25 & 18.37 & 60.11 & 22.12 \\
        & ClipSitu XTF$^*$~\citep{roy2024clipsitu} & {40.01} & {15.03} & {49.78} & \textbf{25.22} & {53.36} & \textbf{33.20} \\
        \rowcolor{gray!20} & Ours & \textbf{41.32} & \textbf{18.16} & \textbf{57.02} & 24.72 & \textbf{65.11} & {27.28} \\
        \hline
    \end{tabular}
    }
    \caption{Full evaluation on the GSR on both validation (dev) and test set of the SWiG dataset~\citep{pratt2020grounded} .
    {\footnotesize $^*$ClipSitu XTF metrics are for the best-performing method presented for verb prediction.}}
\label{tab:gsr_results_app}
\end{table*}

\begin{table*}[t]
\centering
\setlength{\tabcolsep}{5.5pt}
\def\arraystretch{0.95}
\begin{tabular}{lcccccccccccc}
    \toprule
    \multicolumn{1}{c}{} & \multicolumn{4}{c}{WW} & \multicolumn{4}{c}{CC} & \multicolumn{4}{c}{COCO}\\ 
    \cmidrule(lr){2-5} \cmidrule(lr){6-9} \cmidrule(lr){10-13}
    Method  & BL $\uparrow$ & pe $\uparrow$& pc $\downarrow$& sim$\uparrow$ & BL $\uparrow$& pe $\uparrow$& pc$\downarrow$ & sim $\uparrow$& BL $\uparrow$& pe $\uparrow$& pc $\downarrow$& sim $\uparrow$\\ 
    \midrule
    EncDec  & 0.41 & 0.38 & \textbf{0.30} & 0.42 & 0.38 & {0.30} & {0.44} & 0.42 & 0.34 & 0.22 & 0.36 & 0.38 \\ 
    CLIPCap & {0.42} & {0.38} & {0.33} & {0.45} & {0.44} & {0.44} & {0.33} & {0.47} & {0.40} & \textbf{0.40} & {0.30} & {0.47}\\ 
    Ours  & \textbf{0.45} & \textbf{0.43} & {0.33} & \textbf{0.53} & \textbf{0.46} & \textbf{0.51} & \textbf{0.26} & \textbf{0.57} & \textbf{0.43} & \textbf{0.40} & \textbf{0.28} & \textbf{0.52} \\
    
    \midrule

\end{tabular}
\caption{Full HHI results on the Waldo and Wenda benchmark split by data source – Who’s Waldo (WW), Conceptual Captions (CC), and COCO Captions, as described by Alper \etal~\cite{alper2023learning}} 
\label{tab:hhi_results_aa}
\end{table*}

\medskip \noindent
\textbf{Situation Recognition (SiR)}  
 The model is initially trained for 20 epochs with the LoRA configuration  with a rank of 128, an alpha value of 256, and a dropout rate of 0.05. Training optimization is conducted with the AdamW optimizer and learning rate of $1 \times 10^{-4}$ weight decay of 0.01, and an epsilon value of $1 \times 10^{-8}$. This is followed by an additional 7 epochs of fine-tuning with a reduced learning rate of $1 \times 10^{-5}$. We use an effective batch size of 16.

 \medskip \noindent
\textbf{Human-Human Interaction}
  The model is trained for 13 epochs, using LoRA rank of 128, an alpha value of 256, and a dropout rate of 0.05. This employs the AdamW optimizer with a learning rate of $1 \times 10^{-4}$, a weight decay of 0.01, and an epsilon value of $1 \times 10^{-8}$. We use an effective batch size of 16. 

\medskip \noindent
\textbf{Grounded Situation Recognition}  
Building off of CoFormer~\cite{cho2022collaborative}, we apply \attentionmethod to the flattened image features extracted by its CNN backbone (ResNet-50 ~\cite{he2016deep} pretrained on ImageNet ~\cite{deng2009imagenet}), concatenating frozen BLIP-2 features. We train the model using the same parameters recommended in the original method to ensure a fair comparison. We utilize the AdamW optimizer with a weight decay of $10^{-4}$, $\beta_1 = 0.9$, and $\beta_2 = 0.999$. To enhance training stability, gradient clipping is applied with a maximum gradient norm of 0.1. A learning rate scheduler is employed, reducing the learning rates by a factor of 10 at epoch 30. The batch size is set to 16, and the model is trained over 40 epochs.

\medskip \noindent
\textbf{Human-Object Interaction}
Building off of PViC~\cite{zhang2023pvic}, we apply \attentionmethod to the flattened image features extracted by its existing vision backbone, concatenating frozen BLIP-2 features. We use the same hyperparameters as PViC, training for 20 epochs with a 10-fold learning rate reduction at the 10th epoch. We employ the AdamW optimizer, configured with both the learning rate and weight decay set to $10^{-4}$. The model is trained for 30 epochs, with the learning rate reduced by a factor of 5 at the 20th epoch.  All network parameters are fine-tuned, except the (H)-DETR~\citep{carion2020detr,jia2023detrs} object detector backbone, following the original training scheme of PViC.

 \subsection{Models Details}
 
For BLIP-2, we use \texttt{blip2-opt-2.7b} checkpoints\footnote{https://huggingface.co/Salesforce/blip2-opt-2.7b}\footnote{https://huggingface.co/ybelkada/blip2-opt-2.7b-fp16-sharded}. For BLIP, the \texttt{blip-image-captioning-base}\footnote{https://huggingface.co/Salesforce/blip-image-captioning-base} checkpoint; for CLIP (large), the \texttt{clip-vit-lg-p14}\footnote{https://huggingface.co/openai/clip-vit-large-patch14} checkpoint; for CLIP (base), the \texttt{clip-vit-b-p32}\footnote{https://huggingface.co/openai/clip-vit-base-patch32} checkpoint; and for DinoV2, the \texttt{dinov2-base}\footnote{https://huggingface.co/facebook/dinov2-base} checkpoint. 

\section{Additional Task Details}

We proceed to priovide additional details regarding the benchmarks and models used for each task under consideration.

\subsection{Human-Object Interaction Detection (HOI)}
\noindent \textbf{Experimental details}. 
For this task, we utilize the HICO-DET dataset introduced by ~\citep{chao2018learning}, which comprises 37,633 training images and 9,546 test images. This dataset features 80 object classes, using the classes from MS COCO ~\citep{lin2014microsoft}, along with 117 action classes. Altogether, HICO-DET contains 600 fixed categorical HOI classes (\emph{i.e.}, combinations of objects and interactions). 

\medskip \noindent \textbf{Alternative Methods}. We compare performance against PViC, as well as RLIPv2~\citep{yuan2023rlipv2}, that focuses on aligning vision representations with relational texts. In GEN-VLKT~\citep{liao2022gen}, the authors present a method for distilling CLIP ~\citep{radford2021learning} features specifically designed for the human-object interaction (HOI) task.

\subsection{Situation Recognition (SiR)} 
\label{sec:sr}
\medskip \noindent \textbf{Experimental details}. Experiments over this task were trained and evaluated on the imSitu dataset~\citep{yatskar2016situation}. The imSitu dataset contains 75K, 25K and 25K images for train, development and test set, respectively. This dataset contains 504 verbs, 11K nouns and 190 roles.

\medskip \noindent \textbf{Alternative Methods}. 
We compare our framework against ClipSitu ~\citep{roy2024clipsitu} and CoFormer ~\citep{cho2022collaborative}. ClipSitu represents the state of the art, utilizing a cross-attention-based Transformer that employs CLIP visual tokens. CoFormer combines a transformer encoder and decoder to predict both verbs and nouns.

\subsection{Grounded Situation Recognition}
\label{sec:gsr}

\textbf{Experimental details} We use the SWiG dataset ~\citep{pratt2020grounded}, which is an extension of the imSitu dataset ~\citep{yatskar2016situation} that augments its samples with bounding box annotations.

\medskip \noindent 
\textbf{Alternative Methods}. We compare our approach against ClipSitu~\citep{roy2024clipsitu} and CoFormer~\citep{cho2022collaborative}, both of which are also designed for situation recognition. 
\subsection{Human-Human Interaction}
\label{sec:hhi}

\medskip \noindent \textbf{Experimental details}. For this task, we use the Waldo and Wenda benchmark with its accompanying metrics ~\citep{alper2023learning}. The Waldo and Wenda test set contains 1,000 images along with their manually written ground truth HHI labels. These include 238 unique verbs and 575 unique interaction labels. We train on the pseudo-labelled data of Alper \&Averbuch-Elor ~\citep{alper2023learning}, using our Structured Image method with text formatted as HHI de-scriptions with two special token slots for human participants.

\medskip \noindent
\textbf{Alternative Methods} We compare against the SOTA captioning model CLIPCap ~\citep{mokady2021clipcap} and the Vanilla encoder-decoder (EncDec) ~\citep{alper2023learning}, all fine-tuned
with Alper \etal ~\citep{alper2023learning} pseudo-labels for best results. 

\end{appendix}

\end{document}